\pdfoutput=1
\documentclass[runningheads]{llncs}
\usepackage{graphicx}
\usepackage{comment}
\usepackage{amsmath,amssymb} 
\usepackage{color}

\usepackage{algorithm}
\usepackage{algorithmic}
\usepackage{xcolor}
\usepackage{xspace}
\usepackage{multirow}
\usepackage{colortbl}
\usepackage{subfigure}

\usepackage{booktabs}

\newcommand{\cmidr}[1]{\cmidrule(lr){#1}}

\newcommand{\inexactness}{\mathcal{E}}
\newcommand{\flops}{\mathcal{F}}
\DeclareMathOperator*{\argmax}{arg\,max}
\DeclareMathOperator*{\argmin}{arg\,min}
\newcommand{\alphas}{\mathcal{A}}
\newcommand{\invas}{\Theta}
\newcommand{\lag}{\mathcal{L}}
\newcommand{\dsa}{DSA\xspace}

\begin{document}
\pagestyle{headings}
\mainmatter
\def\ECCVSubNumber{922}  



\title{DSA: More Efficient Budgeted Pruning via Differentiable Sparsity Allocation}

\titlerunning{DSA: More Efficient Budgeted Pruning via Differentiable Sparsity Allocation}

\author{Xuefei Ning\inst{1}\thanks{Both authors contributed equally to this work.}\orcidID{0000-0003-2209-8312} \and
Tianchen Zhao\inst{2*}\orcidID{0000-0002-2071-7514} \and
Wenshuo Li\inst{1}\orcidID{0000-0001-5638-2114} \and
Peng Lei\inst{2}\orcidID{0000-0001-7422-0258} \and
Yu Wang\inst{1}\orcidID{0000-0001-6108-5157} \and
Huazhong Yang\inst{1}\orcidID{0000-0003-2421-353X}}

\authorrunning{X. Ning et al.} 
\institute{Department of Electronic Engineering, Tsinghua University \and Department of Electronic Engineering, Beihang University \\
\email{foxdoraame@gmail.com, ztc16@buaa.edu.cn, yu-wang@tsinghua.edu.cn}}
\maketitle

\begin{abstract}

  Budgeted pruning is the problem of pruning under resource constraints. In budgeted pruning, how to distribute the resources across layers (i.e., sparsity allocation) is the key problem. Traditional methods solve it by discretely searching for the layer-wise pruning ratios, which lacks efficiency. In this paper, we propose Differentiable Sparsity Allocation (\dsa), an efficient end-to-end budgeted pruning flow. 
  Utilizing a novel \textit{differentiable pruning process}, DSA finds the layer-wise pruning ratios with \textit{gradient-based optimization}. It allocates sparsity  in continuous space, which is more efficient than methods based on discrete evaluation and search. 
  Furthermore, DSA could work in a \textit{pruning-from-scratch} manner, whereas traditional budgeted pruning methods are applied to pre-trained models. 
  Experimental results on CIFAR-10 and ImageNet show that DSA could achieve superior performance than current iterative budgeted pruning methods, and shorten the time cost of the overall pruning process by at least 1.5$\times$ in the meantime.

  \end{abstract}

  \section{Introduction}
  
  Convolutional Neural Networks (CNNs) have demonstrated superior performances in computer vision tasks. However, CNNs are computational and storage intensive, which poses significant challenges on the NN deployments under resource constrained scenarios. Model compression techniques~\cite{grouplasso,liu2017learning} are proposed to reduce the computational cost of CNNs. Moreover, there are situations (e.g., deploying the model onto certain hardware, meeting real-time constraints) under which the resources (e.g., latency, energy) of the compressed models must be restricted under certain budgets. Budgeted pruning is introduced for handling these situations. 
  
  \begin{figure*}[t]
  \begin{center}
    \includegraphics[width=0.95\linewidth]{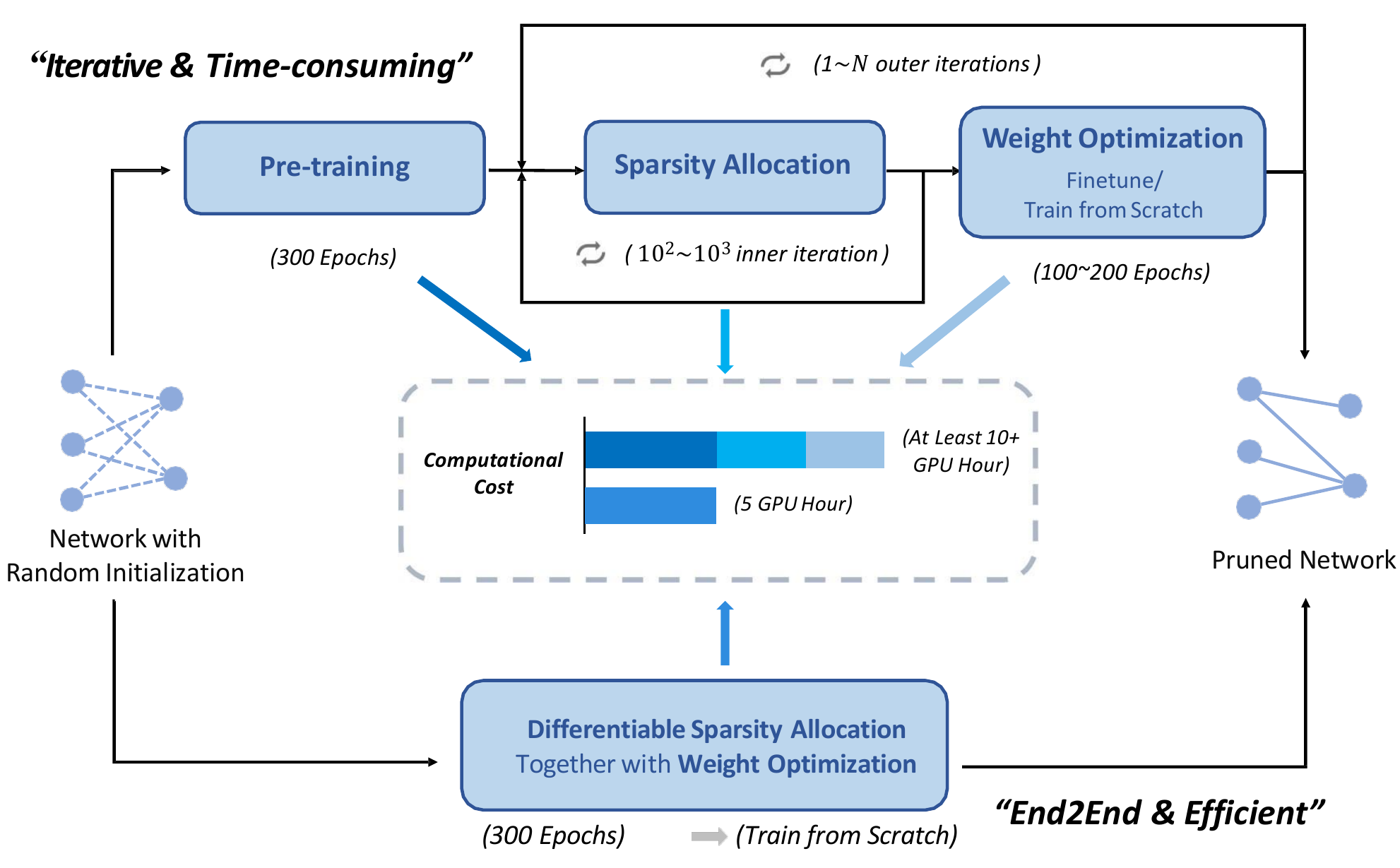}
    \caption{Workflow comparison of the iterative pruning methods~\cite{netadapt,amc,autocompress,liu2019metapruning} and DSA}
    \label{fig:aif_process}
  \end{center}
  \end{figure*}

  As shown in Fig.~\ref{fig:aif_process}, the budgeted pruning problem could be divided into two sub-tasks: to decide how many channels to keep for each layer (i.e., sparsity allocation) and to acquire proper weights (i.e., weight optimization). Recent work~\cite{rethinking} observes that once the pruned structure is acquired, the compressed model can achieve similar accuracies no matter it is trained from scratch or fine-tuned from the weights inherited from the original model. Therefore, sparsity allocation is the key problem for budgeted pruning.
  
  To solve the sparsity allocation problem, the majority of methods~\cite{netadapt,amc,autocompress,liu2019metapruning} adopt an ``iterative pruning flow'' scheme. The workflow of these methods involves three stages: pre-training, sparsity allocation, and finetuning, as shown in Fig.~\ref{fig:aif_process}. These methods conduct the sparsity allocation through a discrete search, which contains hundreds of search-evaluation iterations. For each candidate allocation, a time-consuming approximate evaluation is needed. Also, the discrete search in the large search space lacks sample efficiency. Moreover, these methods need to be applied to the pre-trained models, and model pre-training costs much computational effort. As a result, the overall iterative pruning flow is not efficient.

  In order to improve the efficiency of budgeted pruning, we propose \dsa, an end-to-end pruning flow. 
  Firstly, DSA can work in a ``pruning-from-scratch'' manner, thus eliminating the cumbersome pre-training process (see Sec.~\ref{sec:analysis}). Secondly, \dsa optimizes the sparsity allocation in continuous space with a gradient-based method, which is more efficient than methods based on discrete evaluation and search. 
  
  For applying the gradient-based optimization for allocating sparsity, we should make the evaluation of the validation accuracy and the pruning process differentiable. 
  For the validation performance evaluation, we use the validation loss as a differentiable surrogate of the validation accuracy.
  For the pruning process, we propose a probabilistic differentiable pruning process as a replacement. In the differentiable pruning process, we soften the non-differentiable hard pruning by introducing masks sampled from the probability distributions controlled by the pruning ratio. The differentiable pruning process enables the gradients of the task loss, w.r.t. the pruning ratios to be calculated.
  Utilizing the task loss's gradients, \dsa obtains the sparsity allocation under the budget constraint following the methodology of the Alternating Direction Method of Multipliers (ADMM)~\cite{admm}.
  
  The contributions of this paper are as follows.
  
  \begin{itemize}
  \item \dsa uses \textit{gradient-based optimization} for sparsity allocation under budget constraints, and works in a \textit{pruning-from-scratch} manner. \dsa is more efficient than iterative pruning methods.
  
  \item We propose a novel \textit{differentiable pruning process}, which enables the gradients of the task loss w.r.t. the pruning ratios to be calculated. The gradient magnitudes align well with the layer-wise sensitivity,
    thus providing an efficient way of measuring the sensitivity (See Sec.~\ref{sec:rational}). Due to this property of the gradients, \dsa can attribute higher keep ratios to more sensitive layers.
  
  \item We give a topological grouping procedure to handle the topological constraints that are introduced by skip connections and so on, thus the resulting model keeps the original connectivity.
  
  \item Experimental results on CIFAR-10 and ImageNet demonstrate the effectiveness of \dsa. DSA consistently outperforms other iterative budgeted pruning methods with at least 1.5$\times$ speedup.
   
  \end{itemize}

  \section{Related Work}
  \label{sec:related}
  \subsection{Structured Pruning}
  
  Structured pruning intends to introduce structured sparsity into the NN models. 
  SSL~\cite{grouplasso} removes structured subsets of weights by applying group lasso regularization and magnitude pruning. Some studies~\cite{liu2017learning,morphnet} add regularization on the batch normalization (BN) scaling parameters $\gamma$ instead of the convolution weights. 
  These methods focus on seeking the trade-off between model sparsity and performance via designing regularization terms. Since these methods allocate sparsity through regularizing the weights, the results are sensitive to the hyperparameters.

  There are also studies that target at choosing which filters to prune, given the layer-wise pruning ratios. ThiNet~\cite{luo2017thinet} utilizes the information from the next layer to select filters. 
  FPGM~\cite{fpgm} exploits the distances to the geometric median as the importance criterion. These methods focus on exploring intra-layer filter importance criteria, instead of handling the inter-layer sparsity allocation.

  \subsection{Budgeted Pruning}
  
  To amend the regularization based methods for budgeted pruning, MorphNet~\cite{morphnet} alternates between training with $L_1$ regularization and applying a channel multiplier. 
  However, the uniform expansion of width neglects the different sensitivity of different layers and might fail to find the optimal resource allocation strategy under budget. To explicitly control the pruning ratios, ECC~\cite{ecc}
  updates the pruning ratios according to the energy consumption model. The pruning process is modeled as discrete constraints tying the weights and pruning ratios, and this constrained optimization is handled using a proximal projection. In our work, the pruning process is relaxed into a probabilistic and differentiable process, which enables the pruning ratios to be directly instructed by the task loss.
  
  Other methods view the budgeted pruning problem as a discrete search problem, in which the sparsity allocation and finetuning are alternatively conducted for multiple stages. In each stage, a search-evaluation loop is needed to decide the pruning ratios. 
  For the approximate evaluation, a hard pruning procedure and a walk-through of the validation dataset are usually required. 
  As for the search strategy, NetAdapt~\cite{netadapt} empirically adjusts the pruning ratios, while ensuring a certain resource reduction is achieved;
  AMC~\cite{amc} employs reinforcement learning to instruct the learning of a controller, and uses the controller to sample the pruning ratios; AutoCompress~\cite{autocompress} uses simulated annealing to explore in the search space; 
  MetaPruning~\cite{liu2019metapruning} improves the sensitivity analysis by introducing a meta network to generate weights for pruned networks.
  Apart from these methods that search for the layer-wise pruning ratio,  LeGR~\cite{legr} searches for appropriate affine transformation coefficients to calculate the global importance scores of the filters.
  These methods can guarantee that the resulting models meet the budget constraint, but the discrete search process is inefficient and requires a pre-trained model.
  
  In contrast, \dsa (Differentiable Sparsity Allocation) is an end-to-end pruning flow that allocates the inter-layer sparsity with a gradient-based method, which yields better performance and higher efficiency. Moreover, \dsa works in a ``pruning from scratch'' manner, saving the cost of pre-training the model.
  The comparison of various properties across pruning methods is summarized in Table.~\ref{table:related_work}.

  \begin{table*}[bht]
    \centering
    \caption{Comparison of structured pruning methods. \textbf{Headers}: The ``budget control'' column indicates whether the method could ensure the resulting model to satisfy the budget constraint; The ``from scratch'' column indicates whether the method could be applied to random initialized models rather than pre-trained ones; The ``performance instruction'' column describes how the task performance instructs the sparsity allocation, ``indirect'' means that the task performance instructs the sparsity allocation only indirectly through weights (e.g., magnitude pruning); The ``gen. perf.'' column indicates whether the generalization performance guides the pruning process}
    \label{table:related_work}
    \begin{tabular}{c|cccccc}
    \hline
    \multirow{2}{*}{Methods}  & \multirow{2}{*}{\begin{tabular}[c]{@{}c@{}}budget\\control\end{tabular}} & \multirow{2}{*}{\begin{tabular}[c]{@{}c@{}}from\\scratch\end{tabular}} & \multirow{2}{*}{\begin{tabular}[c]{@{}c@{}}end-to-end\end{tabular}} &  \multirow{2}{*}{\begin{tabular}[c]{@{}c@{}}performance\\instruction\end{tabular}} &  \multirow{2}{*}{\begin{tabular}[c]{@{}c@{}}gen. perf.\end{tabular}}  \\
    \\
      \hline
    SSL~\cite{grouplasso}                & \textbf{no}  & yes & yes & indirect             & no  \\
    NetAdapt~\cite{netadapt}             & yes & no  & no  & discrete evaluation  & yes \\
    AMC~\cite{amc}                       & yes & no  & no  & discrete evaluation  & yes \\
    MetaPruning~\cite{liu2019metapruning}& yes & no  & no  & discrete evaluation  & yes \\
    ECC~\cite{ecc}                       & yes & no  & yes & indirect             & no  \\\hline
    Ours                                 & yes & yes & yes & differentiable       & yes \\\hline
    \end{tabular}
  \end{table*}

  \section{Problem Definition}
  \label{sec:pd}
  
  For budgeted pruning, denoting the budget as $B_\flops$, the weight as $W$, the optimization problem of the keep ratios $\alphas = \{\alpha^{(k)}\}_{k=1,\cdots,K}$ (1 - pruning ratio) of $K$ layers can be written as:
  \begin{equation}
    \begin{aligned}
      \alphas^{*} &= \argmax_{\alphas} \mbox{Acc}_v(W^*(\alphas), \alphas)\\
      \mbox{s.t.}\quad & W^*(\alphas) = \argmin_{W} L_t(W, \alphas)\\
      & \flops(\alphas) \leq B_\flops, \quad 0 \leq \alphas \leq 1
    \end{aligned}
    \label{eq:pd}
  \end{equation}
  where $\mbox{Acc}_v$ is the validation accuracy, and $L_t$ is the training loss, and $\flops(\alphas)$ is the consumed resource corresponding to the keep ratios $\alphas$.
  
  To solve this optimization problem, existing iterative methods~\cite{netadapt,amc,autocompress} conduct sensitive analysis in each stage, and use discrete search methods to adjust $\alphas$. The common assumption adopted is that in each stage, $\mbox{Acc}_v(\hat{W^*}(\alphas), \alphas)$ should be correlated to $\mbox{Acc}_v(W^*(\alphas), \alphas)$, in which $\hat{W^*}(\alphas)$ is approximated using the current weights (e.g., threshold-based pruning, local layer-wise least-square fitting), instead of finding $W^*(\alphas)$ by finetuning. 
  
  \begin{figure*}[t]
  \begin{center}
    \includegraphics[width=0.98\linewidth]{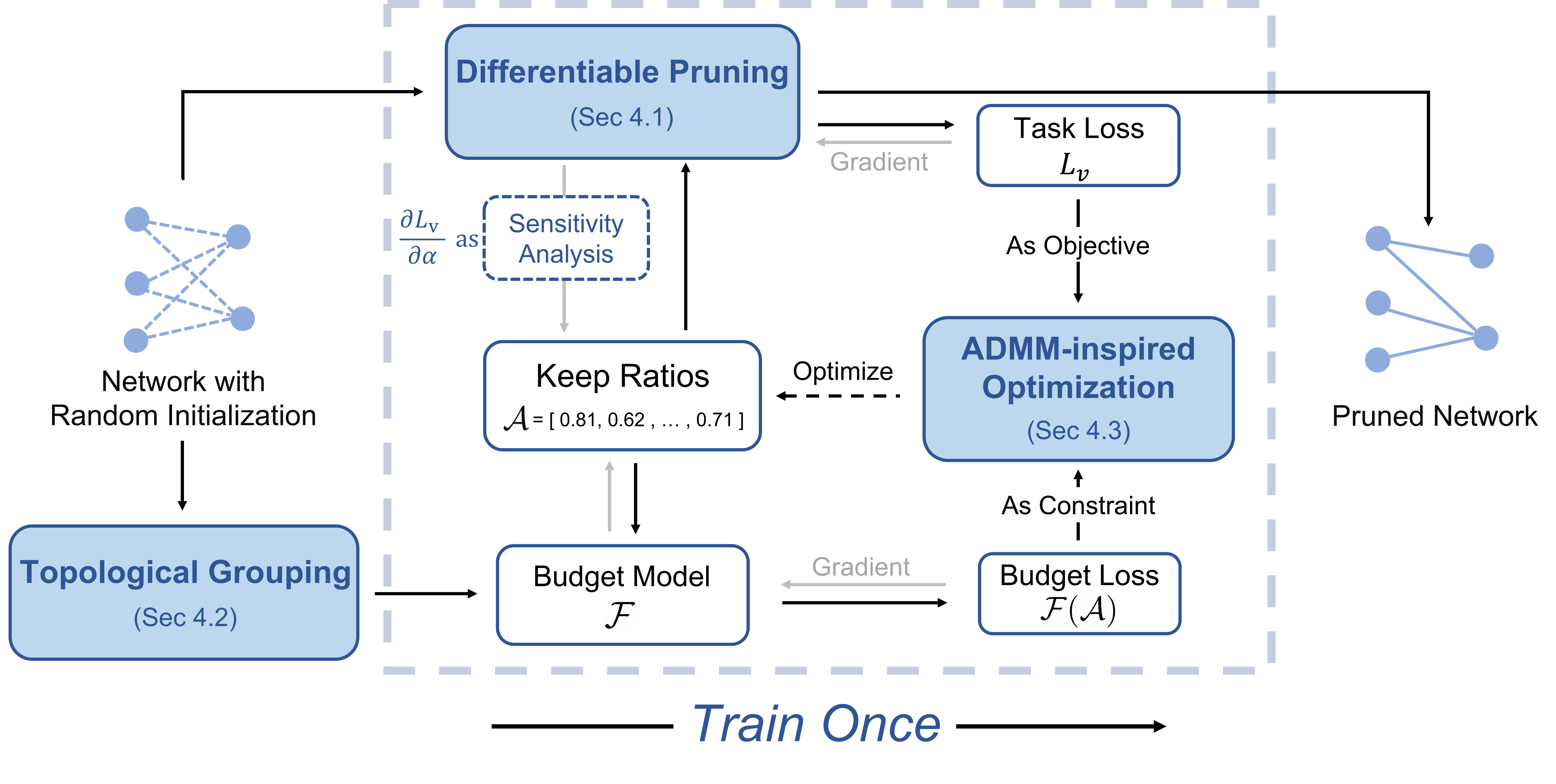}
  \caption{\dsa workflow. For feasible deployment, we first group the network layers according to the topological constraints (Sec.~\ref{sec:method-grouping}), and a keep ratio will be attributed to each topological group.
  The budget model $\flops$ is also generated for measuring the budget loss $\flops(\alphas)$. 
  Then, in the sparsity allocation procedure, the weights are updated using the task loss on the training dataset. And the keep ratios decision (i.e., inter-layer sparsity allocation) is conducted on the validation dataset using gradient-based steps (Sec.~\ref{sec:admm_for_bp}). 
  Note that the differentiable characteristic of the pruning process (Sec.~\ref{sec:method-pruning}) enables the task loss's gradients to flow back to the keep ratios $\alphas$}  
  \label{fig:dsa_process}
  \end{center}
  \end{figure*}

  \section{Method}
  \label{sec:method}

  Since the validation accuracy $\mbox{Acc}_v$ in Eq.~\ref{eq:pd} is not differentiable, we use the validation loss $L_v$ as a differentiable surrogate of $\mbox{Acc}_v$. Then, the objective function becomes $\alphas^{*} = \argmin_{\alphas} L_v(W^*(\alphas), \alphas)$ in the differentiable relaxation of the budgeted pruning problem in Eq.~\ref{eq:pd}.
  As for the inner optimization of $W^*(\alphas)$, we adapt the weights to the changes of the structure by adopting similar bi-level optimization as in DARTS~\cite{darts}. The high-order gradients $\frac{\partial W^*(\alphas)}{\partial \alphas}$ are ignored, thus
  $\frac{\partial L_v(W^*(\alphas), \alphas)}{\partial \alphas} \approx  \frac{\partial L_v(W, \alphas)}{\partial \alphas}$.
  
  The workflow of \dsa is shown in Alg.~\ref{alg:opt} and Fig.~\ref{fig:dsa_process}. 
  First, \dsa groups the network layers according to the topological constraints (Sec.~\ref{sec:method-grouping}), and assigned a keep ratio for each group of layers. The sparsity allocation problem is to decide the proper keep ratios $\alphas$ for the $K$ groups.
  The optimization of the keep ratios $\alphas$ is conducted with gradient-based method in the \emph{continuous} space. 
  We apply an ADMM-inspired optimization method (Sec.~\ref{sec:admm_for_bp}) to utilize the gradients of both the task and budget loss to find a good sparsity allocation $\alpha^*$ that meets the budget constraint.
  Note that to enable the task loss's gradients to flow back to the keep ratios $\alpha$, we design a novel differentiable pruning process (Sec.~\ref{sec:method-pruning}).

  \begin{algorithm}[t]
    \begin{algorithmic}[1]
      \STATE Run topological grouping, get $K$ topological groups, and the budget model $\flops$
    \STATE $\alphas = {\bf 1}_{K}$
  \WHILE{$\flops(\alphas) > B_{\flops}$}
  \STATE Update the keep ratios $\alphas$, auxiliary and dual variables following Eq.~\ref{eq:admm} in Sec.~\ref{sec:admm_for_bp} 
  \STATE Update weights $W$ with SGD:\\ $W_{T} = W_{T-1} - \eta_w \frac{\partial L_t}{\partial W}|_{W_T}$
  \ENDWHILE
  \RETURN the pruned network, $\alphas$
  \end{algorithmic}
  \caption{\dsa: Differentiable sparsity allocation}
  \label{alg:opt}
  \end{algorithm}

  \subsection{Differentiable Pruning}
  \label{sec:method-pruning}
  
  \subsubsection{Pruning Process Relaxation}
  
\begin{figure}[t]
  \begin{center}
    \includegraphics[width=1.0\linewidth]{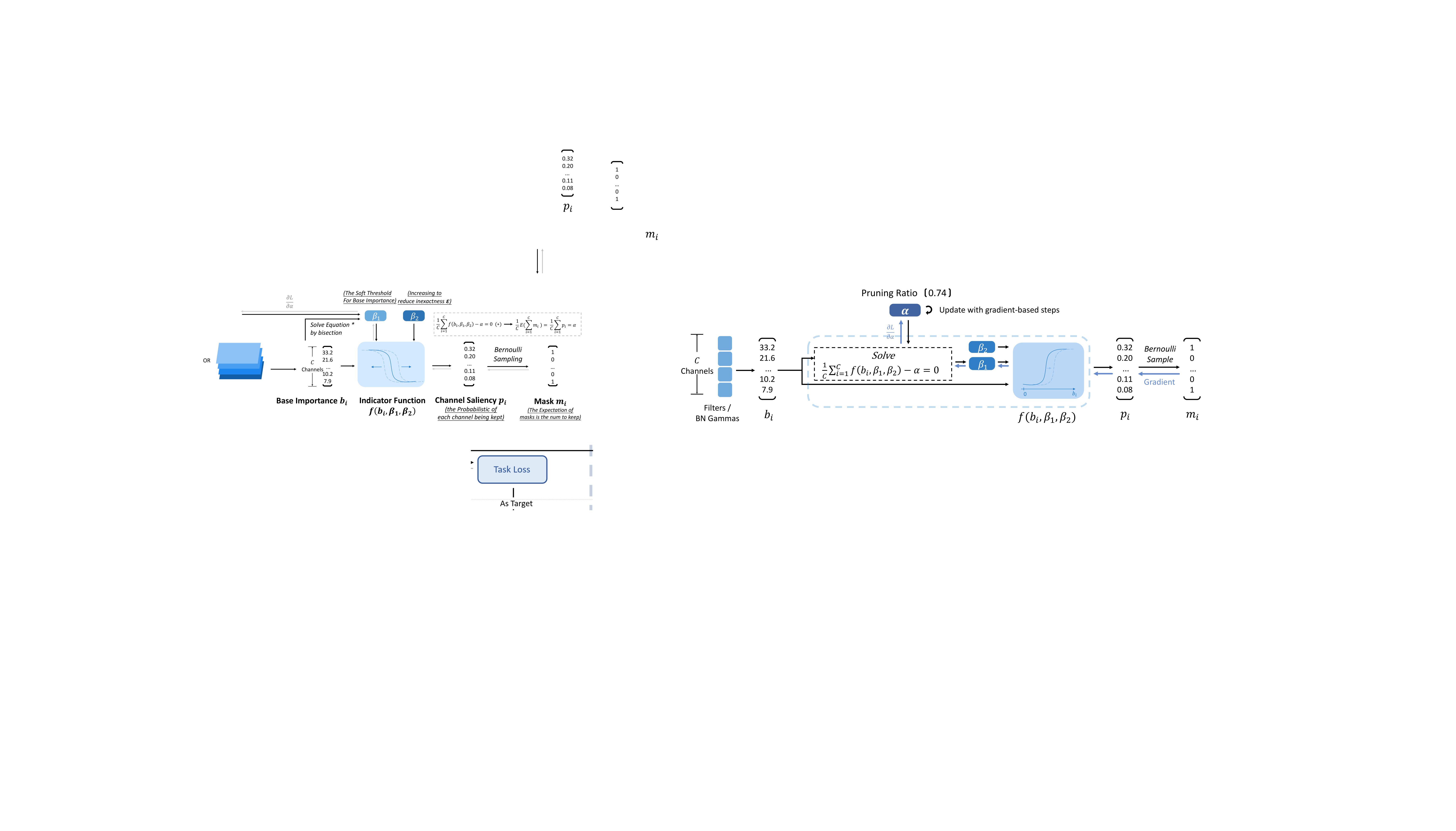}
    \caption{The illustration of the differentiable pruning process of one layer. Given the base importances $b_i$ and the keep ratio $\alpha$, the process outputs channel-wise keep probabilities $p_i = f(b_i, \beta_1, \beta_2)$, which satisfy that the expectation condition $\sum^C_i p_i = \alpha C$. Then, the channel-wise masks $m_i$ are sampled}
  
  \label{fig:differentiable_pruning}
  \end{center}
  \end{figure}

  In the traditional pruning process of a particular layer, given the keep ratio $\alpha$, a subset of channels are chosen according to the channel-wise importance criteria $b_i \in \mathbb{R}^+, i=1,\cdots,C$ (e.g., the L1 norm of the convolutional weights).
  In contrast, we use a probabilistic relaxation of the ``hard'' pruning process where channel $i$ has the probability $p_i$ to be kept.
  Then, channel-wise masks are sampled from the Bernoulli distribution of $p_i$: $m_i \sim \mbox{Bernoulli}(p_i), i=1,\cdots,C$. The pruning process is illustrated in Fig.~\ref{fig:differentiable_pruning}.
  
  The channel-wise keep probability $p_i = f(\alpha, b_i)$ is computed using $\alpha$. 
  Due to the probabilistic characteristics of the pruning process, the proportion of the actual kept channels might deviate from $\alpha$. We should make the expectation of the actual kept channels $E[\sum_{i=1}^C m_i] = \sum_{i=1}^C p_i =\sum_{i=1}^C f(\alpha, b_i)$ equal to $\alpha C$, which we denote as the ``expectation condition'' requirement.
  
  What is more, as we need a ``hard'' pruning scheme eventually, this probabilistic pruning process should become deterministic in the end. Defining the inexactness as $\inexactness = E[|\sum_i m_i - \alpha C|^2] = \sum_i \mbox{Var}[m_i] = \sum_i p_i (1-p_i)$, 
  a proper control on the inexactness is desired such that the inexactness $\inexactness$ can reach $0$ at the end of pruning. 
  
  The choice of $f$ is important to the stability and controllability of the pruning process. An S-shaped function family w.r.t. $b$, $f(b;\beta): \mathbb{R}^+ \to (0, 1)$, parametrized by at least two parameters is required, so that we can control the inexactness and satisfy the expectation condition at the same time.
  We choose the sigmoid-log function $f(b_i, \beta_1, \beta_2) = \mbox{Sigmoid} \circ \mbox{Log}(b_i) =  \frac{1}{1 + (\frac{b_i}{\beta_1})^{-\beta_2}}, \quad \beta_1, \beta_2 > 0$. This function family has the desired property that, 
  $\beta_1$ and $\beta_2$ could be used to control the expected keep ratio $E[\sum_{i=1}^C m_i]$ and the inexactness $\inexactness$ in a relatively independent manner. 1) In our work, $\beta_2$ is a parameter that follows an increasing schedule. As $\beta_2$ approaches infinity, the inexactness $\inexactness$ approaches 0, and $\beta_1$ becomes the hard pruning threshold of this layer.
  2) $\beta_1 = \beta_1(\alpha)$ is a function of $\alpha$. It has the interpretation of the soft threshold for the base importance score. It is calculated by solving the implicit equation of expectation condition: 
  \begin{equation}
      g(\beta_1) = \frac{1}{C} E[\sum_{i=1}^C m_i] - \alpha = \frac{1}{C} \sum_{i=1}^{C} f(b_i, \beta_1, \beta_2) - \alpha = 0
      \label{eq:beta1_eq}
    \end{equation}
    
  Since $g(\beta_1)$ is monotonically decreasing, the root $\beta_1(\alpha)$ could be simply and efficiently found (e.g., with bisection or Newton methods).
  
  To summarize, utilizing the differentiable pruning process, the forward process of the $k$-th layer can be written as
  \begin{equation}
      \begin{aligned}
      & y^{(k)}(w, y^{(k-1)}; \alpha) = m \odot \mbox{Conv-BN-ReLU}(w, y^{(k-1)})\\
      \mbox{s.t.}\quad& m_i \sim \mbox{Bernoulli}(p_i), \quad\sum_{i=1}^C p_i = \sum_{i=1}^Cf(\alpha, b_i) =\alpha C, \quad i=1,\cdots,C
      \end{aligned}
      \label{eq:forward_process}
  \end{equation}
  where $y^{(k-1)}, w, y^{(k)}$ denote the inputs, weights, outputs of this layer, and the superscript $k$ is omitted for simplicity.

  \subsubsection{Differentiable Instruction by the Task Loss}
  
  The task loss $L$ can be written as
  \begin{equation}
      \begin{aligned}
      L(\alphas, W) = E_{x \sim D} [E_{m^{(k)}_i \sim Ber(m; p^{(k)}_i)}[\mbox{CE}(x, y; M, W)]]
      \end{aligned}
  \end{equation}
  where $D$ is the training dataset, $W$ denotes the weights, $M$ is the set of masks $\{\{m^{(k)}_i\}_{i=1,\cdots,C^{(k)}}\}_{k=1,\cdots, K}$, $\alphas$ is the set of keep ratios $\{\alpha^{(k)}\}_{k=1,\cdots,K}$. 
  
  To enable differentiable tuning of the layer-wise keep ratios $\alphas$ instructed by both the task loss and the budget constraint,
  the major challenge is to derive the task loss's gradients w.r.t. $\alpha^{(k)}$: $\frac{\partial L}{\partial \alpha^{(k)}}$. First, we can calculate the implicit gradient $\frac{\partial \beta_1(\alpha)}{\partial \alpha}$ as:
  \begin{equation}
    \begin{aligned}
      &\frac{1}{C} \sum_{i=1}^{C} \frac{\partial f(b_i, \beta_1, \beta_2)}{\partial \beta_1} \frac{\partial \beta_1}{\partial \alpha} - 1 = 0\\
      &\frac{\partial \beta_1(\alpha)}{\partial \alpha}  = \frac{C}{\sum^{C}_{i=1}f'(b_i, \beta_1, \beta_2)}
    \end{aligned}
    \label{eq:grad_beta1_alpha}
  \end{equation}
  
  Then, $\frac{\partial L}{\partial \alpha^{(k)}}$ could be calculated as:
  \begin{equation}
    \begin{aligned}
      \frac{\partial L}{\partial \alpha^{(k)}} &= \frac{\partial \beta_1}{\partial \alpha} \sum_{i=1}^{C} \frac{\partial L}{\partial p_i} \frac{\partial p_i}{\partial \beta_1} = C \sum_{i=1}^{C} \frac{\partial L}{\partial p_i}  \hat{f'_i};\quad \hat{f'_i} = \frac{f'_i}{\sum f'_i}
    \end{aligned}
    \label{eq:grad_L_alpha}
  \end{equation}
  where the superscript $k$ is omitted for simplicity and $\frac{\partial L}{\partial p_i}$ could be approximated using Monte-Carlo samples of the reparametrization gradients. 
    
  Eq.~\ref{eq:grad_L_alpha} could be interpreted as: 
  The update of $\alpha^{(k)}$ is instructed using a weighted aggregation of the gradients of the task loss $L$ w.r.t. the keep probabilities of channels $\frac{\partial L}{\partial p^{(k)}_i}$, and the aggregation weights are $f'_i, i=1,\cdots,C^{(k)}$.

  \subsection{Topological Grouping and Budget Modeling}
  \label{sec:method-grouping}

  For plain CNN, we can choose the layer-wise keep ratios $\alpha^{(k)}, k=1,\cdots,K$ independently. However, for networks with shortcuts (e.g., ResNet), the naive scheme can lead to irregular computation patterns. 
  Our grouping procedure for handling the topological constraints is described in Alg. 1 in the appendix. An example of grouping convolutions in two residual blocks is also shown in the appendix.
  
  The $\flops(\alphas)$ function models relationship of the keep ratios $\alphas$ and the resources. Taking FLOPs as an example, $\flops(\alphas)$ could be represented as $\flops(\alphas) = \alphas^T \flops_A \alphas + \flops_B^T \alphas$. 
  For completeness, we summarize the calculation procedure of $\flops_A$ and $\flops_B$ in Alg.~1 in the appendix. 
  Under budget constraints for resources other than FLOPs, regression models can be fitted to get the corresponding $\flops$ model.


  \subsection{ADMM-inspired Method For Budgeted Pruning}
  \label{sec:admm_for_bp}

  ADMM is an iterative algorithm framework that is widely used to solve unconstrained or constrained convex optimization problems. Here,
  we use alternative gradient steps inspired by the methodology of ADMM to solve the constrained non-convex optimization problem.

  Substituting the variable $\alphas$ by $\invas = \mbox{Sigmoid}^{-1}(\alphas)$,
  the $0 \leq \alphas \leq 1$ constraints are satisfied naturally. 
  By introducing auxiliary variable $z$ and the corresponding dual variable $u_2$ for the equality constraint $z = \invas$, the augmented Lagrangian is:
  \begin{equation}
      \begin{aligned}
      \lag(\invas, z, u_2) &= L_v(\invas) + I(\flops(z) \leq B_\flops) + u_2^T (\invas - z) + \frac{\rho_2}{2} ||\invas - z||^2
      \end{aligned}
  \end{equation}
  
  We then minimize the dual lower bound $\max_{u_2} \lag(\invas, z, u_2)$ of $L_v(\invas)$.
  Eq.~\ref{eq:admm} shows the 3 alternative steps in one iteration. 
  The variables with the superscript ``$'$'' denote the values at the previous time step.
  
  \begin{equation}
      \begin{aligned}
      \invas &= \argmin_\invas L_v(\invas) +  u_2^T (\invas - z') + \frac{\rho_2}{2} ||\invas - z'||^2\\
      z &= \argmin_z u_2^T (\invas - z) + \frac{\rho_2}{2} ||\invas - z||^2 \quad\mbox{s.t.} \quad \flops(z) \leq B_\flops\\
      u_2 &= u_2' + \rho_2 (\invas - z)
      \end{aligned}
  \label{eq:admm}
  \end{equation}

  The unconstrained sub-problem for $\invas$ is hard to solve, since $L_v(\invas)$ is a stochastic objective and $W$ can only be regarded as being constant in a local region. Therefore, in each iteration, we only do one stochastic gradient step on one validation batch for $\invas$. 
  
  To solve the inner problem for the auxiliary variable $z$ with an inequality constraint, we use the standard trick of converting $\flops(z) \leq B_\flops$ to $[\flops(z) - B_\flops]_+ = 0$, and then use gradient descent to solve the min-max optimization of the augmented lagrangian $\lag^{(z)}(z, u_1)$ in Eq.~\ref{eq:inner_z_opt}. In each iteration of the inner optimization, one gradient descent step is applied to $z$: $z = z' - \eta_z \nabla_z\lag^{(z)}(z, u_1)$, and one dual ascent step is applied to $u_1$: $u_1 = u_1' + \rho_1 [\flops(\invas) - B_\flops]_+$. This optimization is efficient since only $z$ and $u_1$ need to be updated.

  \begin{equation}
  \lag^{(z)}(z, u_1) = u_1 [\flops(z) - B_\flops]_+ \frac{\rho_1}{2} [\flops(z) - B_\flops]^2 + u_2^T (\invas - z) + \frac{\rho_2}{2} ||\invas - z||^2
  \label{eq:inner_z_opt}
  \end{equation}

  The dual variables $u_1, u_2$ can be interpreted as the regularization coefficients, which are dynamically adjusted according to the constraint violations.

    \begin{table}[H]
    \caption{Pruning results of ResNet-20 and ResNet-56 on CIFAR-10. SSL and MorphNet are re-implemented with topological grouping. Accuracy drops for the referred results are calculated based on the reported baseline in their papers. \textbf{Headers}: ``TG'' stands for Topological Grouping; ``FLOPs Budget'' stands for the percentage of the pruned models' FLOPs compared to the full model}
    \label{tab:res_cifar10}
    \begin{center}
    \resizebox{1.0\textwidth}{!}{
    \begin{tabular}{ccccccc}
    \toprule
    \multirow{3}{*}{\begin{tabular}[c]{@{}c@{}}FLOPs\\ Budget\end{tabular}} & \multirow{3}{*}{Method} &  \multirow{3}{*}{\begin{tabular}[c]{@{}c@{}}TG\end{tabular}} & \multicolumn{2}{c}{ResNet-20} & \multicolumn{2}{c}{ResNet-56}\\ \cmidrule(lr){4-5} \cmidrule(lr){6-7} 
    &  & & \begin{tabular}[c]{@{}c@{}}FLOPs\\ ratio\end{tabular} & Acc. (Acc. Drop)    & \begin{tabular}[c]{@{}c@{}}FLOPs\\ ratio\end{tabular} & Acc. (Acc. Drop)    \\ \midrule
    \multicolumn{1}{c|}{\multirow{1}{*}{\begin{tabular}[c]{@{}c@{}} Baseline\end{tabular}}}   & \multicolumn{2}{c}{Ours} & 100 \%  & 92.17 \% & 100 \% & 93.12 \% \\ \cmidr{1-7}
    \multicolumn{1}{c|}{\multirow{4}{*}{\begin{tabular}[c]{@{}c@{}}75\% \end{tabular}}}   & SSL~\cite{grouplasso} & $\checkmark$ & 73.8\%  & 91.08\% (-1.09\%)  & 69.0\% & 92.06\% (-1.06\%) \\
    \multicolumn{1}{c|}{} & Variational$^*$~\cite{variational}& &  83.5\%  & 91.66\% (-0.41\%) & 79.7\% & 92.26\% (-0.78\%) \\
    \multicolumn{1}{c|}{} & PFEC$^*$~\cite{pfec} &  \checkmark & -  & -  & 74.4\% & 91.31\% (-1.75\%)       \\
    \multicolumn{1}{c|}{} & MorphNet~\cite{morphnet} & $\checkmark$& 74.9\%  & 90.64\% (-1.53\%) & 69.2\% & 91.71\% (-1.41\%)  \\\cmidr{2-7}
    \multicolumn{1}{c|}{} & DSA (Ours)            & $\checkmark$ & 74.0\%  & \textbf{92.10\% (-0.07\%)} & 70.7\% & \textbf{93.08\% (-0.04\%)}  \\ \cmidr{1-7}
    \multicolumn{1}{c|}{\multirow{8}{*}{50\% ($\times 2$)}} 
                          & SSL~\cite{grouplasso} &$\checkmark$  & 51.8\%$^\dagger$  & 89.78 \% (-2.39\%) & 45.5\% & 91.22\% (-1.90\%) \\
    \multicolumn{1}{c|}{} & MorphNet~\cite{morphnet} &  $\checkmark$   & 47.1\%  & 90.1\% (-2.07\%) & 51.9\%$^\dagger$  & 91.55\% (-1.57\%) \\
    \multicolumn{1}{c|}{} & AMC$^*$~\cite{amc}  &  $\checkmark$  & - & - & 50\%   & 91.9\% (-0.9\%) \\
    \multicolumn{1}{c|}{} & CP$^*$~\cite{cp}    &    & - & - & 50\%   & 91.8\% (-1.0\%) \\
    \multicolumn{1}{c|}{} & Rethink$^*$~\cite{rethinking}  & & 60.0\% & 91.07\% (-1.34\%) & 50\%  & 93.07\% (-0.73\%) \\
    \multicolumn{1}{c|}{} & SFP$^*$~\cite{sfp}  &    & 57.8\%  & 90.83\% (-1.37\%) & 47.4\% & 92.26\% (-1.33\%)       \\
    \multicolumn{1}{c|}{} & FPGM$^*$~\cite{fpgm}&    & 57.8\%   & 91.09\% (-1.11\%) & 47.4\%  & \textbf{92.93\%} (-0.66\%) \\
    \multicolumn{1}{c|}{} & LCCL$^*$~\cite{LCCL}&   & 64.0\%   & \textbf{91.68\%} (-1.06\%) & 62.1\%  & 92.81\% (-1.54\%) \\
    \cmidr{2-7}
    \multicolumn{1}{c|}{} & DSA (Ours) &  $\checkmark$   & 49.7\%   & 91.38\% \textbf{(-0.79\%)} & 47.8\%  & 92.91\% \textbf{(-0.22\%)} \\ \cmidr{1-7}
    \multicolumn{1}{c|}{\multirow{3}{*}{\begin{tabular}[c]{@{}c@{}}33.3\% ($\times 3)$\end{tabular}}} & SSL~\cite{grouplasso} &  $\checkmark$ & 34.6\%$^\dagger$ & 89.06\% (-3.11\%) & 38.1\%$^\dagger$ & 91.32\% (-1.80\%)  \\
    \multicolumn{1}{c|}{} & MorphNet~\cite{morphnet} &  $\checkmark$ & 30.5\%  & 88.72\% (-3.45\%) & 39.7\%$^\dagger$ & 91.21\% (-1.91\%) \\\cmidr{2-7}
    \multicolumn{1}{c|}{} & DSA (Ours)           &  $\checkmark$    & 32.5\%  & \textbf{90.24\% (-1.93\%)} & 32.6\% & \textbf{92.20\% (-0.92\%)} \\ \bottomrule
    \end{tabular}
    }
    \begin{minipage}{0.98\textwidth}
      $\dagger$: These pruned models' FLOPs of SSL and MorphNet are higher than the budget constraints, since
      these regularization based methods lack explicit control of the resource consumption and even with carefully tuned hyperparameters, the resulting model might still violate the budget constraint. 
      
      $^*$: These methods' results are directly taken from their paper, and their accuracy drops are calculated based on their reported baseline accuracies.
    \end{minipage}
    \end{center}
    \end{table}

  \begin{table}[bt]
    \begin{center}
      \caption{Pruning results on ImageNet. ``TG'' stands for Topological Grouping}
      \label{tab:res_imgnet}
      \begin{tabular}{c|ccccc}
        \toprule
        \specialrule{0em}{1pt}{5pt}
        \multirow{2}{*}{Network} & \multirow{2}{*}{TG} & \multirow{2}{*}{Method} & FLOPs & Top-1 & Top-5 \\ 
                                 & & & Ratio & Acc Drop & Acc Drop \\
        \cmidrule{1-6} 
        \multirow{5}{*}{ResNet18   } &  & Baseline & 100\%  &  \textbf{69.72}\% & \textbf{89.07\%}\\
        \cmidrule{2-6}  
                                 & & MiL~\cite{LCCL}  & 65.4\%  &  -3.65\% & -2.30\% \\  
                                 & & SFP~\cite{sfp} & 60.0\%  &  -3.18\% & -1.85\% \\  
                                 & & FPGM~\cite{fpgm} & 60.0\%  &  -2.47\% & -1.52\% \\ 
        \cmidrule{2-6}   
                                 & \checkmark & Ours & 60.0\%  &  \textbf{-1.11\%} & \textbf{-0.718\%} \\ 
        \cmidrule{1-6}
        \multirow{12}{*}{ResNet50   } &  & Baseline & 100\%  &  \textbf{76.02\%} & \textbf{92.86\%}\\
        \cmidrule{2-6} 
                                 & & APG~\cite{APG} & 69.0\%  &  -1.94\% & -1.95\% \\  
                                 & & GDP~\cite{GDP} & 60.0\%  &  -2.52\% & -1.25\% \\  
                                 & & SFP~\cite{sfp} & 60.0\%  &  -1.54\% & -0.81\% \\ 
                                 & & FPGM~\cite{fpgm} & 60.0\%  &  -1.12\% & -0.47\% \\ 
        \cmidrule{2-6}
                                 & \checkmark& Ours & 60.0\%  &  \textbf{-0.92\%} & \textbf{-0.41\%} \\ 
        \cmidrule{2-6}
                                 & & ThiNet~\cite{luo2017thinet}  & 50.0\%  &  -4.13\% & - \\ 
                                 & & CP~\cite{cp}  & 50.0\%  &  -3.25\% & -1.40\% \\ 
                                 & & FPGM~\cite{fpgm}  & 50.0\%  &  -2.02\% & -0.93\% \\ 
                                 & & PFS~\cite{PFS}  & 50.0\%  &  -1.60\% & - \\ 
                                 & & Hinge~\cite{hinge} & 46.55\% & -1.33\% & - \\
        \cmidrule{2-6}
                                 & \checkmark & Ours & 50.0\%  &  \textbf{-1.33\%} & \textbf{-0.8\%} \\ 
        \bottomrule
      \end{tabular}
    \end{center}
  
  \end{table}

  \section{Experiments}
  \label{sec:exp}
  \subsection{Setup}

  
  We conduct the experiments on CIFAR-10 and ImageNet. 
  For CIFAR-10, the batch size is 128, and an SGD optimizer with momentum 0.9, weight decay 4e-5 is used to train the model for 300 epochs. The learning rate is initialized to 0.05 
  and decayed by 10 at epochs 120, 180, and 240. The differentiable sparsity allocation is conducted simultaneously with normal training after 20 epochs of warmup. 
  As for ImageNet, we use an SGD optimizer (weight decay 4e-5, batch size 256) to optimize the models for 120 epochs. The learning rate is 0.1 and decayed by 10 at epochs 50, 80, 110. The first 15 epochs remains plain training without pruning. 
  
  In the sparsity allocation process, 10\% of the training data are used as the validation data for updating the keep ratios, while 90\% are used for tuning the weights. For the optimization of $\alphas$, the penalty parameters $\rho_1, \rho_2$ is set to 0.01, and the scaling coefficient of $L_v(\invas)$ is 1e+5. $z$ is updated for 50 steps with learning rate 1e-3 in the inner optimization. In practice, to reach the budget faster, we project the gradients of $\invas$ to be nonnegative. After each update of $\alphas$, the weights are tuned for 20 steps for adaption. After acquiring the budget, the whole training set is used for updating the weights. 
  
  In the differentiable pruning process, the $L_1$ norms of BN scales are chosen as the base importance scores $b_i =|\gamma_i|$. $\beta_1(\alpha)$ is found by solving Eq.~\ref{eq:beta1_eq} with the bisection method. 
  $\beta_2(T)$ follows a increasing schedule: starts at 0.05 and gets multiplied by 1.1 every epoch. As $\beta_2 \to \infty$, the soft pruning process becomes a hard pruning process, and the inexactness $\inexactness$ goes to $0$.
  
  \subsection{Results on CIFAR-10 and ImageNet}
  
  On CIFAR-10,
  for SSL~\cite{grouplasso} and MorphNet~\cite{morphnet}, the regularization coefficients on the convolution weights or BN scaling parameters are adjusted to meet various budget constraints.

\begin{figure*}[ht]
    \centering
  \subfigure[ResNet-20]{\includegraphics[width=0.48\linewidth]{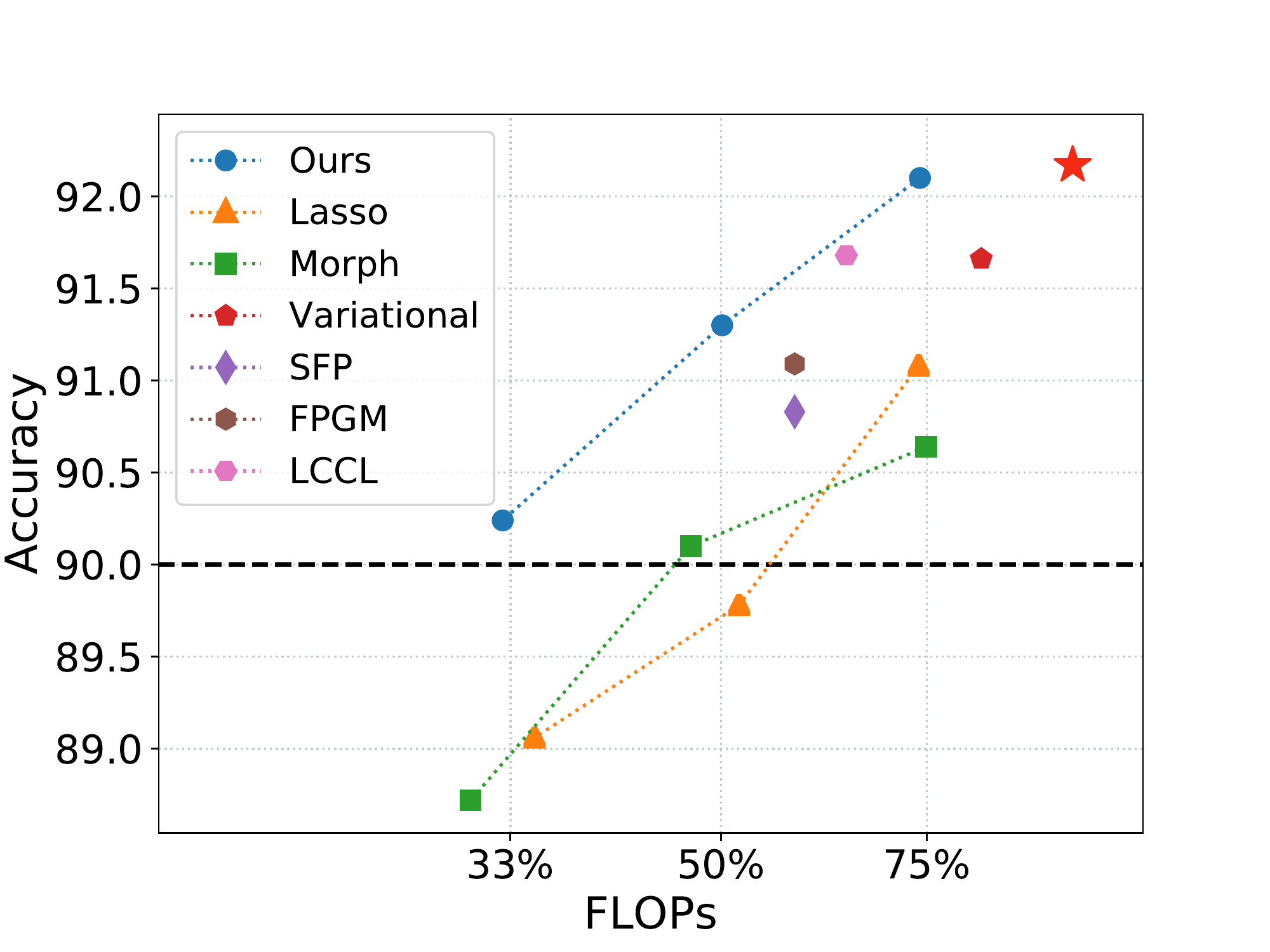}}
  \subfigure[ResNet-56]{\includegraphics[width=0.48\linewidth]{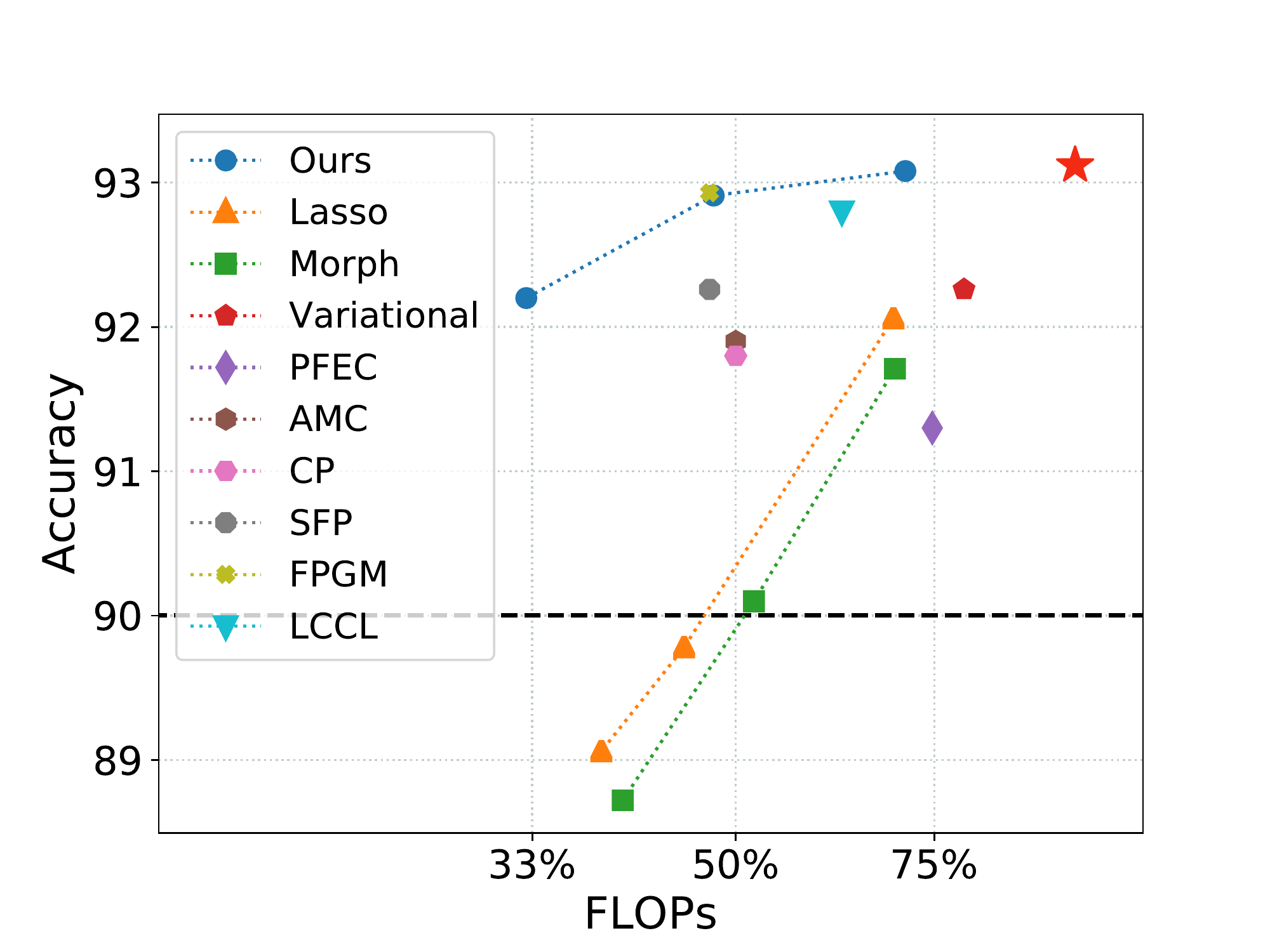}}
  \caption{Pruning results on CIFAR-10}
  \label{fig:results_cifar10}
  \end{figure*}

  Table.~\ref{tab:res_cifar10} and Fig.~\ref{fig:results_cifar10} show the results of pruning ResNet-20 and ResNet-56 on CIFAR-10. The pruned models obtained by \dsa meet the budget constraints with smaller accuracy drops than the baseline methods. 
  Compared with the regularization based methods (e.g., SSL and MorphNet), due to the explicit budget modeling, \dsa guarantees that the resulting models meet different budget constraints, without hyperparameter trials. Compared with the iterative pruning methods (e.g., AMC), DSA allocates the sparsity in a gradient-based way and is more efficient (See Sec.~\ref{sec:analysis}).
  We also apply \dsa to compress ResNet-18 and VGG-16, and the results are included in Appendix Table. 1 and Fig. 2. It shows that DSA outperforms the recent work based on a discrete search~\cite{autocompress}. For ResNet-18, \dsa achieves 94.19\% versus 93.91\% of the baseline with roughly the same FLOPs ratio. For VGG-16, \dsa achieves 90.16\% with 20.4$\times$ FLOPs reduction, which is significantly better than the baseline~\cite{autocompress} (88.78\% with $14.0\times$ FLOPs reduction).

  Table~\ref{tab:res_imgnet} shows the results of applying DSA to prune ResNet-18 and ResNet-50 on ImageNet. 
  As could be seen, DSA consistently outperforms other methods across different FLOPs ratios and network structure. For example, \dsa could achieve a small accuracy drop of 1.11\% while keeping 60\% FLOPs of ResNet-18, which is significantly better than the baselines. 

  \subsection{Analysis and Discussion}
  \label{sec:analysis}
  
  \begin{figure*}[ht]
    \begin{center}
      \subfigure[Sensitivity analysis.]{\includegraphics[width=0.45\linewidth]{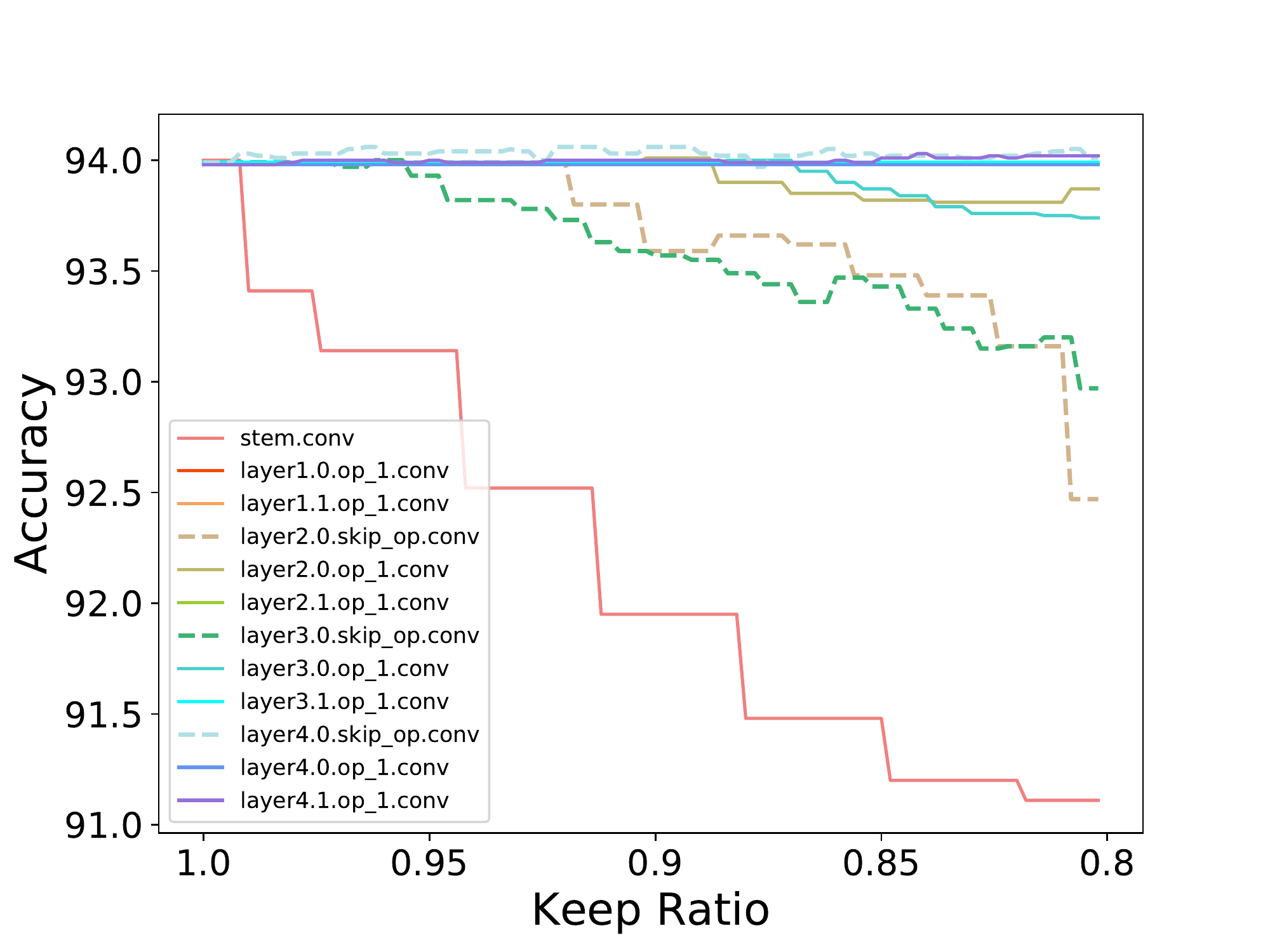}\label{fig:sens}}
      \subfigure[The normalized magnitudes for layer-wise sensitivity and task loss's gradients in the first epoch.]{\includegraphics[width=0.45\linewidth]{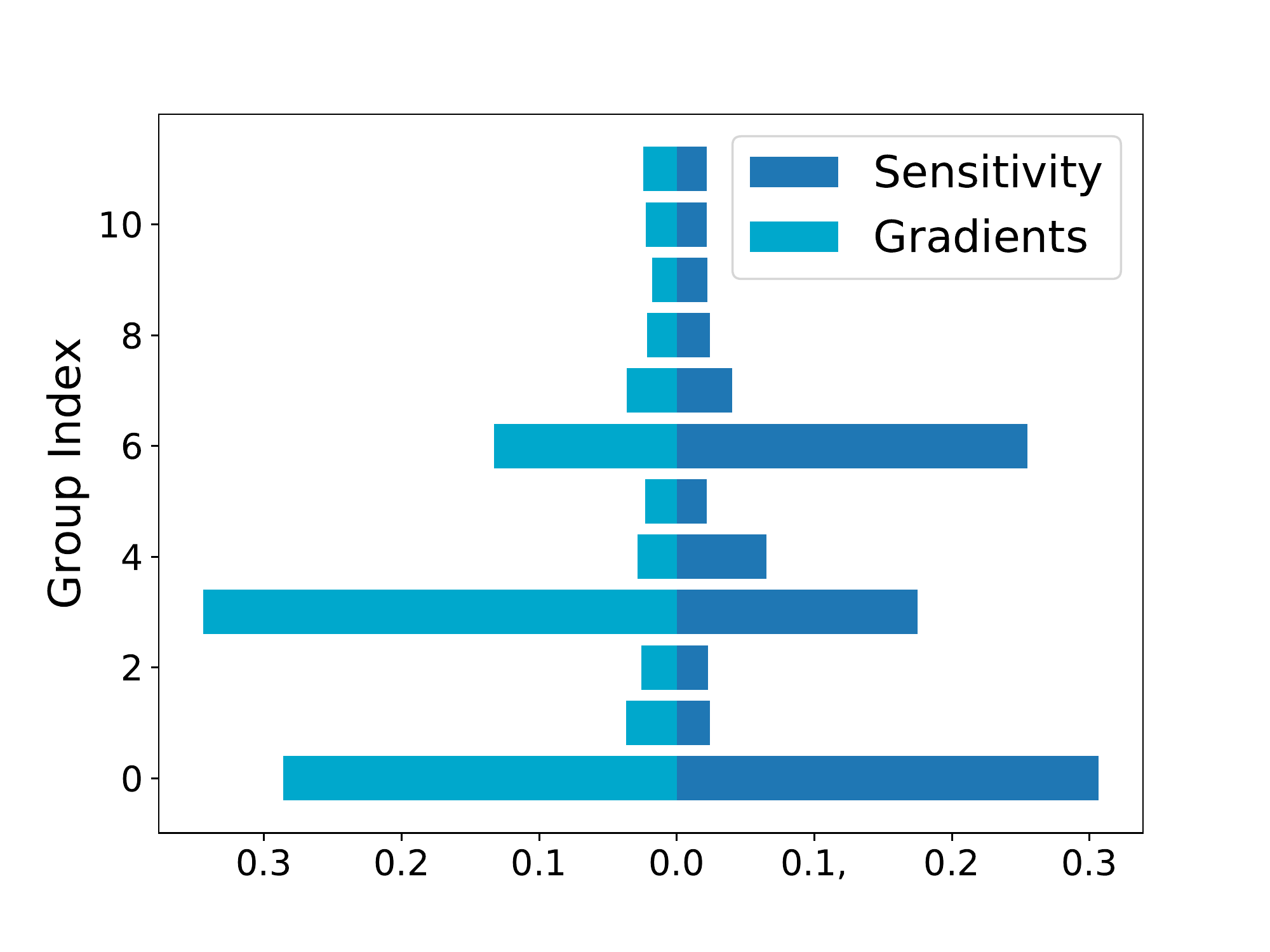}\label{fig:sens_grad}}
      \caption{The alignment between sensitivity analysis and gradient magnitudes of ResNet-18 on CIFAR-10. The magnitudes are normalized by $\hat{v}=\mbox{softmax}(v/\mbox{std}(v))$}
      \label{fig:sens_res18}
    \end{center}
  \end{figure*}

  \subsubsection{Computational Efficiency}

  
  Some recent studies~\cite{rethinking,PFS} suggest that starting with a pre-trained model might not be necessary for pruning. Unlike current methods which rely on a pre-trained model, \dsa could work in a ``pruning from scratch'' manner. 

  As shown in Fig.~\ref{fig:aif_process}, traditional budgeted pruning consists of 3 stages: pre-training, sparsity allocation, and finetuning. 
  The iterative pruning methods conduct hundreds of search-evaluation steps for sparsity allocation, e.g., AMC takes about 3 GPU hours for pruning ResNet-56 on CIFAR-10~\cite{PFS}. After learning the structure, the finetuning stage takes 100\~{}200 epochs for CIFAR-10 (60 epochs for ImageNet), which accounts for about 2\~{}3 GPU hours for ResNet-56 on CIFAR-10 and 150 GPU hours for ResNet-18 on ImageNet. Moreover, these two stages should be repeated for multiple rounds to achieve the maximum pruning rates~\cite{netadapt,autocompress}, thus further increase the computational costs by several times. 
  What's more, these methods need to be applied to the pre-trained models, and the pre-training stage takes about 300 and 120 epochs for models on CIFAR-10 and ImageNet. To summarize, the 3 stages can take up to 10 GPU hours for ResNet-56 on CIFAR-10, and 450 GPU hours for ResNet-18 on ImageNet. 
  In contrast, the sparsity allocation in DSA is carried out in a more efficient gradient-based way, without the need of the pre-trained models. The extra cost of the sparsity allocation is small, since all the ADMM updates can be merged into the optimization of weights, and are conducted only once every tens of weight optimization steps. 
  The whole DSA flow runs for 300 and 120 epochs on CIFAR-10 and ImageNet (5/300 GPU hours), thus speed up the overall pruning process by about 1.5$\times$.

  \subsubsection{Rationality of the Differentiable Sparsity Allocation}
  \label{sec:rational}
  
  In DSA, the task loss's gradient w.r.t. layer-wise pruning ratios directly guides the budget allocation. To see whether the gradient magnitudes align well with the local sensitivity, we conduct an empirical sensitivity analysis for ResNet-18 on CIFAR-10. We prune each layer (topological group) independently with different pruning ratios according to the $L_1$ norm, and show the test accuracy in Fig.~\ref{fig:sens}. Although this sensitivity analysis is heuristic and approximate, the accuracy drop could be interpreted as the local sensitivity for each group. Fig.~\ref{fig:sens_grad} shows the normalized magnitudes of the task loss's gradients and the sensitivity of the layer-wise sparsity. We can see that these two entities align well, giving evidence that the task loss's gradient indeed encodes the relative layer-wise importance.

\begin{figure}[hb]
    \begin{center}
      \includegraphics[width=0.9\linewidth]{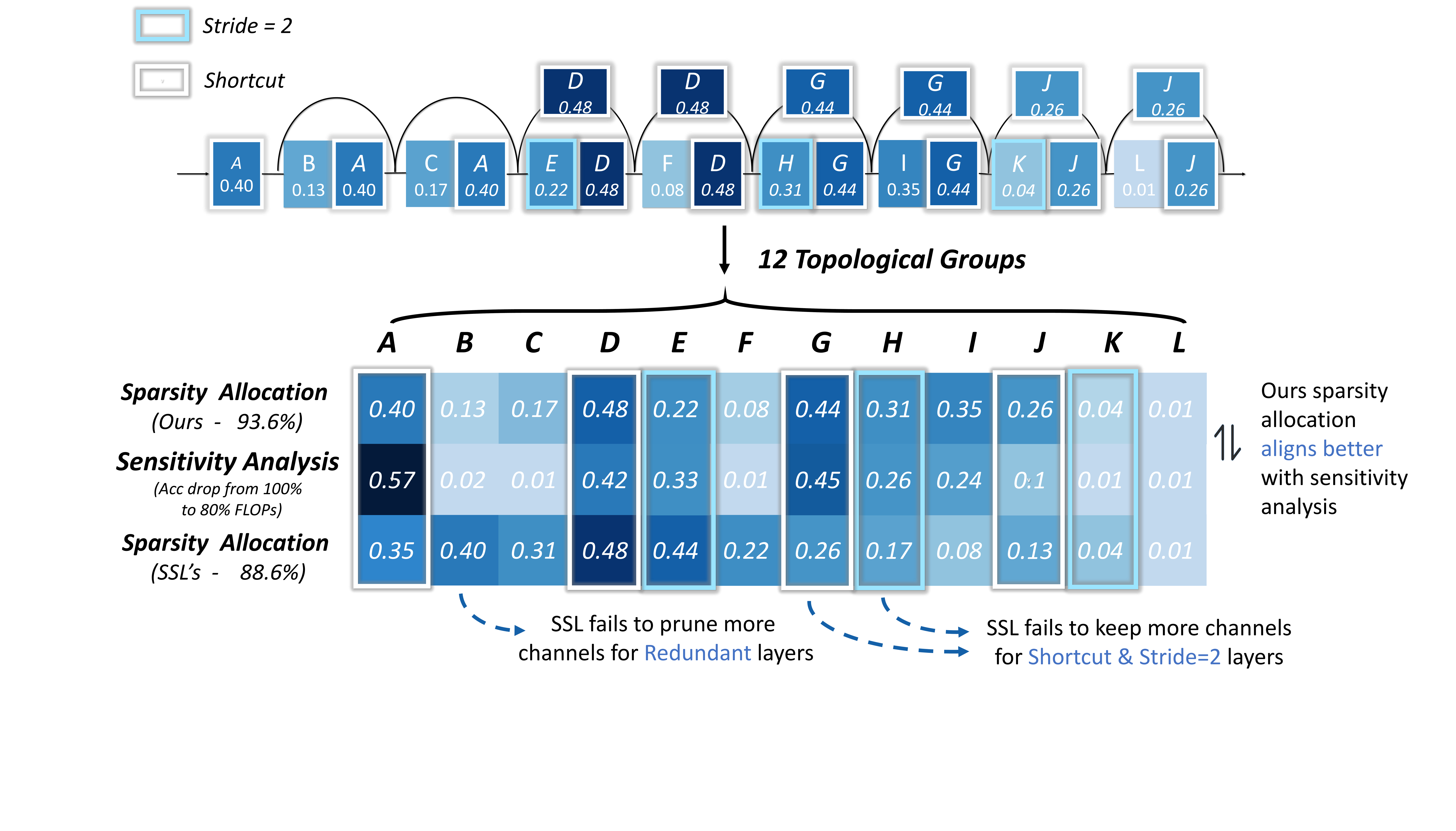}
      \caption{The comparison between the normalized sensitivity, and the sparsity allocation of \dsa and SSL~\cite{grouplasso} for ResNet-18 on CIFAR-10 }
      \label{fig:heat}
    \end{center}
  \end{figure}

  Fig.~\ref{fig:heat} presents the sparsity allocation (FLOPs budget 25\%) for ResNet-18 on CIFAR-10 obtained by \dsa and SSL~\cite{grouplasso}.
  The results show that the first layer for primary feature extraction should not be pruned too aggressively (group A), 
  so do shortcut layers that are responsible for information transmission across stages (groups A, D, G, J). The strided convolutions are relatively more sensitive, 
  and more channels should be kept (groups E, H). 
  In conclusion, DSA obtains reasonable sparsity allocation that matches empirical knowledge~\cite{liu2019metapruning,rethinking}, with lower computational cost than iterative pruning methods.

  \section{Conclusion}
  \label{sec:conclusion}
  In this paper, we propose Differentiable Sparsity Allocation (DSA), a more efficient method for budgeted pruning. Unlike traditional discrete search methods, DSA optimizes the sparsity allocation in a \textit{gradient-based way}.  To enable the gradient-based sparsity allocation, we propose a novel \textit{differentiable pruning process}.
  Experimental results show that DSA could achieve superior performance than iterative pruning methods, with significantly lower training costs.
  
  \section*{Acknowledgments}
  This work was supported by National Natural Science Foundation of China (No. 61832007, 61622403, 61621091, U19B2019), 
  Beijing National Research Center for Information Science and Technology (BNRist). The authors thank Novauto for the support.

  \clearpage

  \bibliographystyle{splncs04}
  \bibliography{egbib}{}
  
  \clearpage

  \pagestyle{headings}
  \title{Appendices for DSA: More Efficient Budgeted Pruning via Differentiable Sparsity Allocation}
  \author{}
  \institute{}
  \titlerunning{Appendix}
  \authorrunning{X. Ning et al.}
  
  \maketitle
  
  \section{Topological Grouping and Budget Model $\flops$}
  
\begin{figure}[h]
  \begin{center}
  \includegraphics[width=0.85\linewidth]{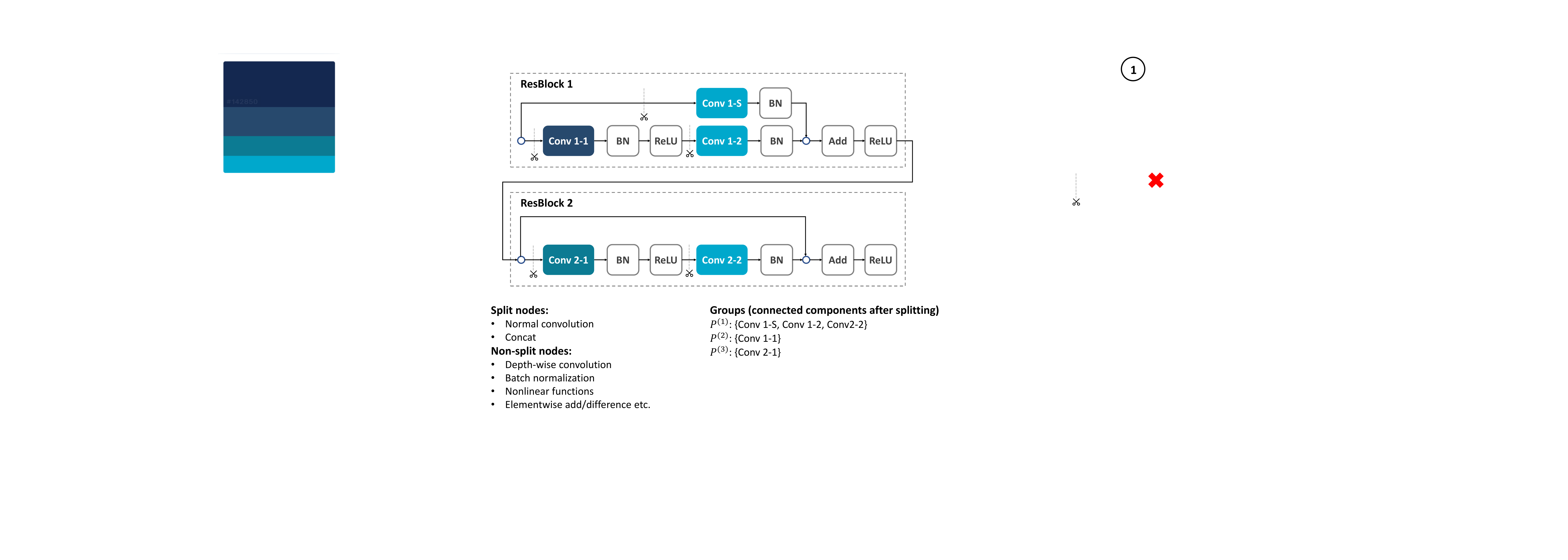}
  \caption{An example of the topological grouping procedure}
  \label{fig:grouping}
  \end{center}
  \end{figure}

  An example of grouping convolutions in two consecutive residual blocks is shown in Fig.~\ref{fig:grouping}. All the incoming connections of the normal convolutions are removed, and then the convolutions in each connected component belong to the same topological group (i.e., share the same keep ratio $\alpha^{(k)}$ and masks ${m^{(k)}_{i=1,\cdots,C}}$).
  
  \begin{algorithm}[h]
  \begin{algorithmic}[1]
  \STATE Construct the computational directed acyclic graph $G$
  \STATE Removing all the incoming connections at split nodes (operation nodes that is not a channel-wise operation): normal convolution, concat operation. \\NOTE: Non-split nodes include all channel-wise operations: depthwise convolution, element-wise add, ReLU, batch normalization, etc.
  \STATE Find the connected components $\{P^{(k)}\}_{k=1,\cdots,K}$. All the convolution layers in each $P^{(k)}$ share the same keep ratio $\alpha^{(k)}$ and masks ${m^{(k)}_{i=1,\cdots,C}}$
  \end{algorithmic}
  \caption{Topological grouping procedure}
  \label{alg:grouping}
  \end{algorithm}

  A budget model $\flops$ is needed for measuring the resource consumption $\flops(\alphas)$ corresponding to the sparsity allocation $\alphas$. Taking FLOPs as an example, $\flops(\alphas)$ could be represented as $\flops(\alphas) = \alphas^T \flops_A \alphas + \flops_B^T \alphas$. We summarize the calculation procedure of $\flops_A$ and $\flops_B$ in Alg.~\ref{alg:flops}.

  \begin{algorithm}[H]
  \begin{algorithmic}[1]
  \STATE $G$: the directed acyclic graph of operations
  \STATE $K$: number of connected components
  \STATE $\flops_A = {\bf 0}_{K \times K}, \flops_B={\bf 0}_{K}$
  \STATE Convolution node attributes: 1) $C$: output channel number; 2) $k$: the index of the connected components that the convolution node belongs to; 3) kss: kernel spatial size (e.g. $3\times 3 = 9$); 4) oss: output spatial size (e.g. $16\times 16=256$)
  \FORALL{convolution node $M$ in $G$}
  \IF{n.type == DEPTHWISE\_CONV}
  \STATE $\flops_B[M.k] += 2 \times M.c \times M.\mbox{kss} \times M.\mbox{oss}$
  \ELSE
      \STATE stack = [predecessor($M$)]
      \WHILE{stack}
      \STATE n = stack.pop()
      \IF{n.type == CONCAT}
      \FOR{pn in predecessor(n)}
      \STATE stack.push(pn)
      \ENDFOR
      \ELSIF{n.type in ELEMENTWISE\_OPs (e.g. ADD, ReLU)}
      \STATE stack.push(predecessor(n)[0])
      \ELSIF{n.type == NORMAL\_CONV}
      \STATE $\flops_A[M.k, n.k] += 2 \times n.C \times M.C \times M.\mbox{kss} \times M.\mbox{oss}$
      \ELSIF{n.type == INPUT}
      \STATE $\flops_B[M.k] += 2 \times n.C \times M.C \times M.\mbox{kss} \times M.\mbox{oss}$
      \ELSE
      \STATE stack.push(n)
      \ENDIF
      \ENDWHILE
  \ENDIF
  \ENDFOR
  \RETURN $\flops_A, \flops_B$
  \end{algorithmic}
  \caption{Calculation of $\flops_A, \flops_B$ ($\flops$ for FLOPs resource)}
  \label{alg:flops}
  \end{algorithm}

  \section{Additional Results}
  
\begin{figure*}[h]
    \begin{center}
      \subfigure[ResNet-18]{
        \centering
        \includegraphics[width=0.45\linewidth]{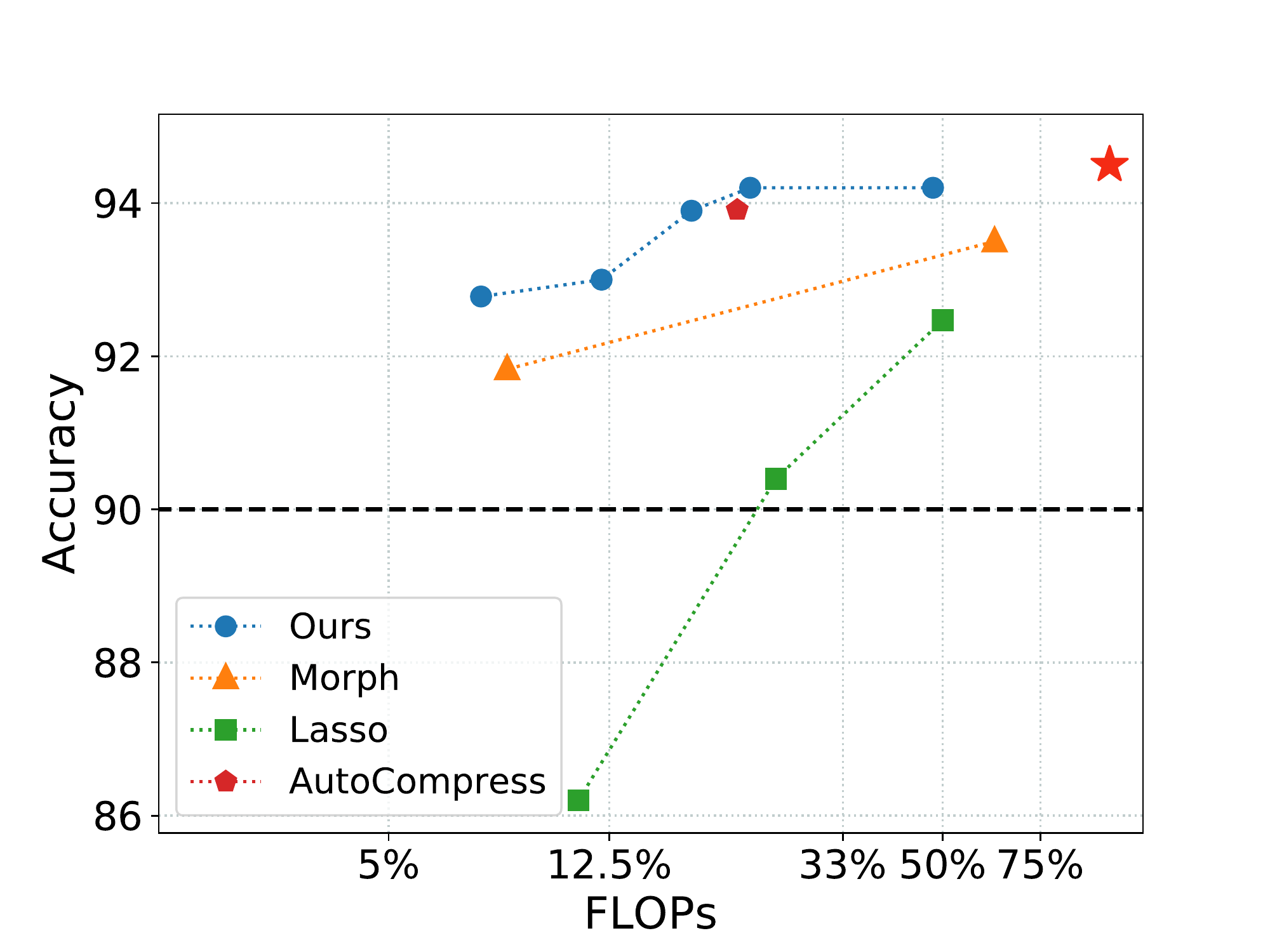}  
        \label{fig:res18_cifar10}
      }
      \subfigure[VGG-16]{
        \centering
        \includegraphics[width=0.45\linewidth]{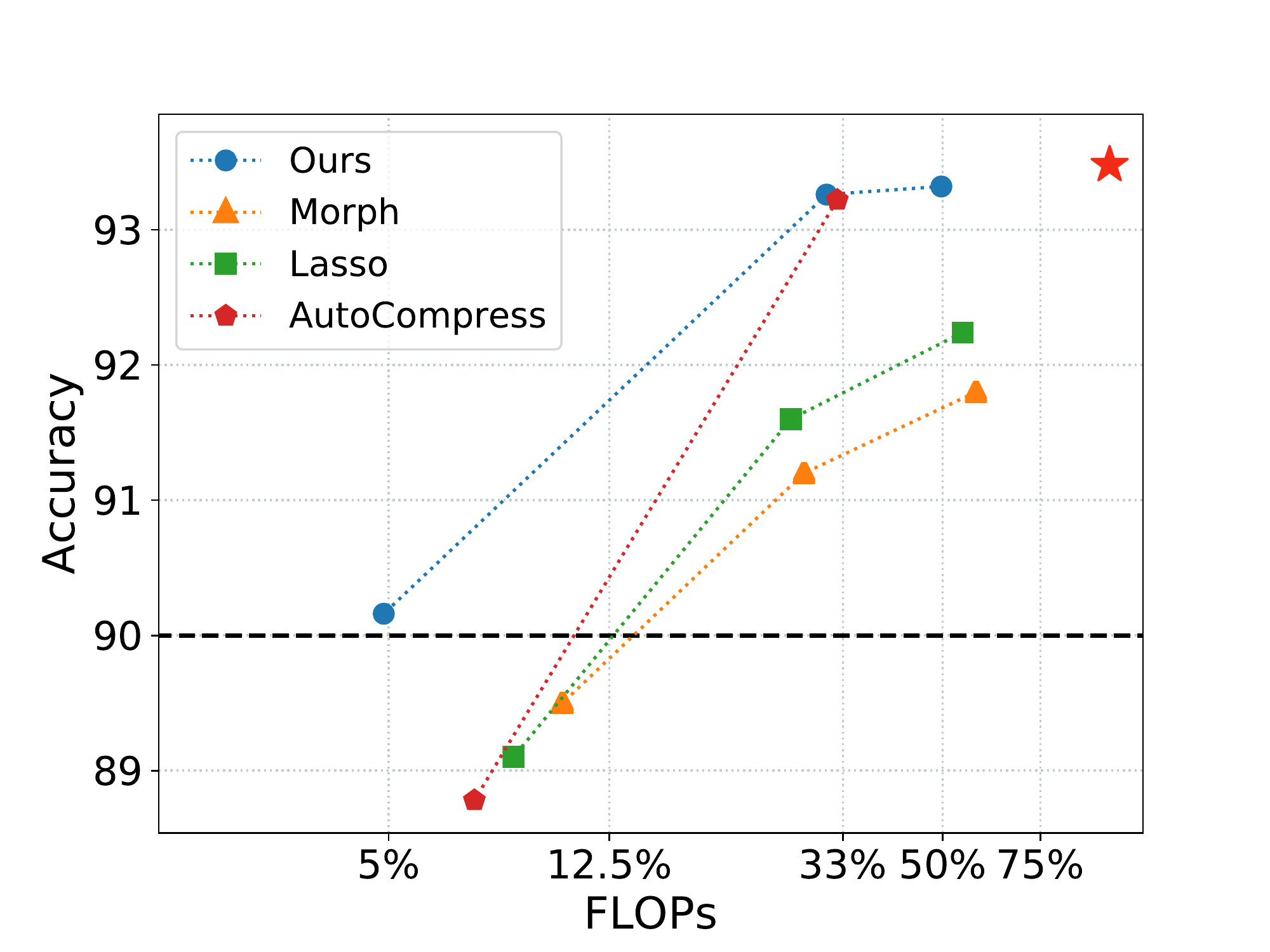}  
        \label{fig:vgg_cifar10}
      }
      \caption{Pruning results of ResNet-18, VGG-16 on CIFAR-10}
      \label{fig:results_cifar10_18vgg}
    \end{center}
  \end{figure*}

  Table.~\ref{tab:res_cifar10_18} and Fig.~\ref{fig:results_cifar10_18vgg} shows the results of pruning ResNet-18 and VGG-16 on CIFAR-10.

  \addtolength{\tabcolsep}{1pt}
  \begin{table}[h]
  \caption{Pruning results of ResNet-18 and VGG-16 on CIFAR-10}
  \label{tab:res_cifar10_18}
  \begin{center}
    \begin{tabular}{c@{\hskip 0.02\linewidth}cccc}
      \toprule
  \multirow{2}{*}{Method} & \multicolumn{2}{c}{ResNet-18 (94.0\%)} & \multicolumn{2}{c}{VGG-16 (93.48\%)} \\\cmidrule(lr){2-3}\cmidrule(lr){4-5} 
                          & Acc.            & FLOPs ratio          & Acc.       & FLOPs ratio      \\\midrule
  \multirow{2}{*}{AutoCompress~\cite{autocompress}}      & 93.91\%         & 4.7$\times$ (21.28\%)        & 93.22\%    &
  3.1$\times$ (32.26\%)    \\
    & -       &  -       & 88.78\%   &  14.0$\times$ (7.14\%) \\\cmidrule(lr){1-5} 
  \multirow{2}{*}{DSA (Ours)}               & {\bf 94.19\%}           & 4.5$\times$ (22.46\%)        &  {\bf 93.26\%}    & 3.2$\times$ (30.85\%)             \\
            & 93.90\%          &  {\bf 5.7$\times$ (17.54\%)}     &  90.16\%    &     {\bf 20.4$\times$ (4.90\%)}   \\\bottomrule
  \end{tabular}
  \end{center}
  \end{table}
  \addtolength{\tabcolsep}{-1pt}
  
  \end{document}